\documentclass[sigconf,nonacm]{acmart}

\usepackage[normalem]{ulem}
\usepackage{multicol}
\usepackage{graphicx}
\usepackage{url}
\usepackage{wrapfig}
\usepackage{balance}

\usepackage{cleveref}
\usepackage{multirow}
\usepackage{subfigure}
\usepackage{latexsym}

\usepackage{mdframed}
\usepackage{array}
\newcolumntype{P}[1]{>{\raggedright\arraybackslash}p{#1}}

\usepackage{microtype}
\usepackage{subfigure}
\AtBeginDocument{%
  \providecommand\BibTeX{{%
    \normalfont B\kern-0.5em{\scshape i\kern-0.25em b}\kern-0.8em\TeX}}}

\copyrightyear{2023} 
\acmYear{2023} 
\setcopyright{rightsretained} 
\acmConference[JCDL '23]{Proceedings of the 23nd ACM/IEEE Joint Conference on Digital Libraries}{June 26--30, 2023}{Santa Fe, New Mexico, USA}
\acmBooktitle{Proceedings of the 23nd ACM/IEEE Joint Conference on Digital Libraries}\acmDOI{10.1145/xxxxx.yyyy}
\acmISBN{978-1-4503-xxx/yy/zz}

\usepackage{booktabs} 
\usepackage{color}
\usepackage{footnote}
\makesavenoteenv{tabular}
\makesavenoteenv{table}

\usepackage{enumitem}
\setlist[itemize,1]{leftmargin=0.4cm}

\begin{document}

\fancyhead{}
\title{Extreme Classification for Answer Type Prediction in Question Answering}

\author{Vinay Setty}
\affiliation{%
\institution{Department of Electrical Engineering and Computer Science}
  \institution{University of Stavanger}
  \city{Stavanger}
  \country{Norway}
}
\email{vsetty@acm.org}

\begin{abstract}
Semantic answer type prediction (SMART) is known to be a useful step towards designing effective question answering (QA) systems. The SMART task involves predicting the top-$k$ knowledge graph (KG) types for a given natural language question. This is challenging due to the large number of types in KGs. In this paper, we propose use of extreme multi-label classification using Transformer models (XBERT) by clustering KG types using structural and semantic features based on question text. We specifically improve the clustering stage of the XBERT pipeline using the  features derived from KGs. We show that these features can improve end-to-end performance for the SMART task, and yield state-of-the-art results. 
\end{abstract}



\maketitle

\section{Introduction}
\label{sec:intro}

Question answering (QA) systems provide concise answers to natural language questions. Predicting the semantic answer type of question is an important component of QA systems~\cite{Moldovan:2003:TOIS,Ferrucci:2010:AI, Kamath:2019:CS,Roy:2021:QABook}.
Answer type prediction in QA is in fact not a recent problem.
Early works have focused on identifying coarse-grained types, e.g., wh-types of questions (who, when, what and where) \cite{Li:2006:NLE,Huang:2008:EMNLP,Kwok:2001:WWW, Riloff:2000:ANLP-NAACL,Voorhees:2001:TREC,Bast:2015:CIKM}. 
Mapping a given question to a semantic type, from type systems of large knowledge graphs (KGs), are known to benefit both open domain QA~\cite{Sun:2015:WWW} and factoid questions in KGQA~\cite{Nikas:2021:ISWC,Yavuz:2016:EMNLP,Shen:2019:EMNLP}. 

Both fine and coarse-grained type prediction are complementary to each other.  Recognizing this, the \emph{semantic answer type prediction} (SMART) challenge~\cite{Mihindukulasooriya:2020:ISWCSMART} was organized as part of the International Semantic Web Conference in 2020,\footnote{\url{https://smart-task.github.io}} introducing the following task: given a natural language question, predict both the high-level answer category and a list of fine-grained types from an underlying KG (DBpedia or Wikidata).
Coarse-grained categories are boolean, literal, or resource; fine-grained types only apply for the resource category.

The most effective approaches to the SMART challenge are based on Transformer models~\cite{Nikas:2020:ISWCSMART,Setty:2020:ISWCSMART}.
Vanilla Transformer models work well out-of-the-box for coarse-grained category prediction, where there are only a handful of possible classes.
However, they cannot be used for classifying the large number of semantic resource types found in KGs; for example, DBpedia has over 760 types and Wikidata has over 50k types~\cite{Chang:2020:KDD}. 
To overcome this, we propose a BERT-based extreme multi-label classification technique (XBERT) has been proposed that solves this problem using a three-stage pipeline: (1) \textbf{Clustering} reduces the number of target classes, (2) \textbf{Classifying} trains the Transformer model to predict type clusters to which the question belongs to, and (3) \textbf{Ranking} selects the top types within the cluster using a linear ranker~\cite{Setty:2020:ISWCSMART}.

In this three stage pipeline, we show that just by leveraging the \emph{clustering} step, we can gain significant performance improvement in answer type classification.
Since clustering is the first step in the three-stage pipeline, any errors occurring during this stage will propagate downstream and thus hinder the entire pipeline's performance.
Instead of naively clustering the KG types based on associated question text as in~\cite{Setty:2020:ISWCSMART}, we aim to leverage additional information from the KG, like type descriptions and entity-type assignments.
We show that using such additional KG features can significantly improve performance, without modifying other stages in the pipeline.

The type systems of KGs can vary a lot in terms of scale and depth of hierarchy~\cite{Garigliotti:2019:IET}, which represents another challenge. We therefore propose a KG-agnostic type clustering technique and perform experiments on two of the most popular KGs, DBpedia and Wikidata, with very different type systems.

Our study is driven by the following research questions:
\begin{itemize}
    \item \textbf{RQ1}: Can we improve the clustering stage of the XBERT pipeline for the SMART task using signals from the KG?
    \item \textbf{RQ2}: How well do these findings generalize across KGs, which differ in the characteristics of their type systems? 
\end{itemize}
The main technical contribution of this paper is the use of signals from a KG to improve the BERT-based extreme multi-label classification approach (XBERT) for the SMART task.
Our experiments show that these features can improve results for both DBpedia and Wikidata and yield state-of-the-art performance over vanilla BERT.

\begin{figure*}[hbt!]
  \centering
  \vspace{-\baselineskip}
  \includegraphics[width=0.8\linewidth]{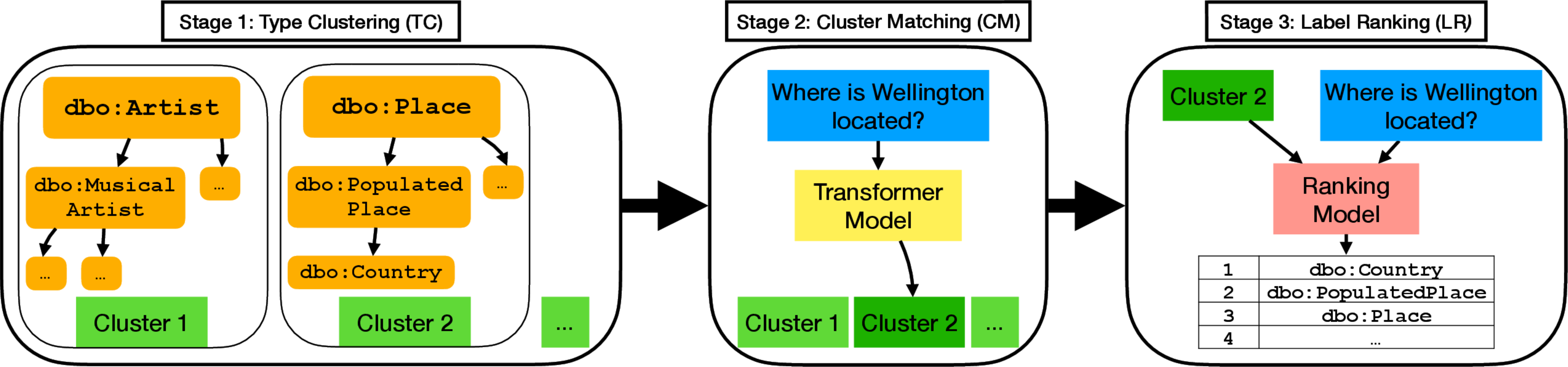}
  \vspace{-\baselineskip}
  \caption{Three phases of semantic answer type classification.}
  \label{fig:detailed}
  
\end{figure*}
\section{Related work}
\label{sec:related}

The correct prediction of the expected answer type is shown to be one of the most important factors to a QA system's overall performance~\cite{Moldovan:2003:TOIS}.
Types may be coarse grained entity classes ~\cite{Nikas:2021:ISWC,Yavuz:2016:EMNLP,Shen:2019:EMNLP} or fine-grained semantic types from the type systems of large knowledge graphs (like Wikidata, DBpedia and Freebase)~\cite{Perevalov:2021:ISWC,Sun:2015:WWW}. 

In \cite{Sun:2015:WWW}, they use probabilistic models to verify if the candidate answer types match the expected answer types to the question. 
Answer type prediction is also related to the task of inferring semantic types of queries, referred to as \emph{target entity type identification}~\cite{Balog:2018:book}, which has been studied in the context of entity-oriented search~\cite{Vallet:2008:SIGIR,Balog:2012:HTT,Garigliotti:2017:SIGIR}.
There, it is approached as a ranking task, using different ways of aggregating entity descriptions~\cite{Balog:2012:HTT}, which can be combined with additional taxonomic and embedding-based features~\cite{Garigliotti:2017:SIGIR}.
Unlike in ad hoc retrieval, the evaluation measure considers the hierarchical relationships between types~\cite{Balog:2012:HTT}---the same methodology has been followed in the SMART challenge.

Most participants at the SMART challenge employed classification methods.  Common themes include data augmentation~\cite{Perevalov:2021:ISWCSMART,Perevalov:2021:ISWC}, the use of word embeddings to represent the questions and types~\cite{Bill:2020:ISWCSMART,Vallurupalli:2020:ISWCSSMART}, while the top performing approaches employed BERT-based classifiers~\cite{Nikas:2020:ISWCSMART,Setty:2020:ISWCSMART}.
Closest to our approach is \cite{Nikas:2020:ISWCSMART} in second place, performing two stage (coarse and fine-grained) type classification. For fine-grained type prediction, they train a BERT-based classifier using the  most frequent types in the dataset. In their follow-up paper~\cite{Nikas:2021:ISWC}, they show how to use answer type prediction to improve a KGQA system. However, since this approach can predict only top-level types, it does not scale to type systems with a large number of types with a relatively flat hierarchy such as Wikidata. Moreover, the performance of this method on DBpedia is also not comparable to the XBERT approach. We extend and improve the top performing solution in \cite{Setty:2020:ISWCSMART}, who introduce the use of XBERT for the SMART challenge.

Several extreme classification approaches have been proposed in the literature~\cite{Prabhu:2018:WWW,Liu:2020:arXiv}. We use XBERT, which provides a good trade-off between performance and efficiency~\cite{Chang:2020:KDD}. A more efficient version of XBERT is also available~\cite{Yu:2020:arXiv}.

\section{Problem Statement}
\label{sec:prob}

The SMART task is defined as follows: given a natural language input question $q$, (1) predict the coarse category $l \in \{\texttt{boolean}, \texttt{number},$ $\texttt{string}, \texttt{date}, \texttt{resource}\}$, and (2) if the category is \texttt{resource}, then return a list of top-$k$ most relevant types $t \in T$ from the type system $T$ of an underlying knowledge graph (KG). 

The first sub-task, \emph{coarse category classification}, may be regarded as a solved problem, as vanilla BERT models can perform it with over 97\% accuracy~\cite{Setty:2020:ISWCSMART}.  
The second sub-task, \emph{type prediction}, proves to be more challenging due to the large number of types in KGs, ranging from several hundreds (e.g., DBpedia) to tens of thousands (e.g., Wikidata). 
The type prediction can be viewed as learning a multi-label classifier, which assigns a score to a given question and type $(q, t)$ pair.

The main challenge lies within the resource category, like \emph{Who are some players of the England national football team?}, where the task is not only about identifying matching types, but also ranking them in order of relevance, from most specific to more generic: \texttt{["dbo:SoccerPlayer", "dbo:Athlete", "dbo:Person", "dbo:Agent"]}. %

\subsection{Solution Overview}

To overcome the limitation of vanilla transformers for large number of classes, we use the XBERT approach by \cite{Chang:2020:KDD}, which uses a three-stage pipeline for the type prediction task (see Fig.~\ref{fig:detailed}).

\begin{description}[leftmargin=0.3cm]
    \item[Type Clustering (\textit{TC}):] First, all unique types $T'\subset T$ which are in the training data are clustered using type vectors. In~\cite{Setty:2020:ISWCSMART}, types are represented using TF-IDF vectors, computed from the question text of all the instances for a given type in the training data. In this paper, we use structural and textual features derived from the KGs to represent types. This stage is the main focus of this paper and it is elaborated further in Sect.~\ref{sec:approach}.
    \item[Cluster Matching (\textit{CM}):] Next, we fine-tune a pre-trained Transformer model to match a given question $q$ into one of the clusters $c$ from the previous stage. The output of this stage is a cluster matching score, $m(q,c)$, which is computed for each cluster based on the model's confidence.
    \item[Label Ranking (\textit{LR}):]  
    Finally, a one-vs-all linear classifier is trained for each type within a cluster $c$ and matched to the given input question $q$ to predict a relevance score $h(q,t)$ for each type in that cluster. The final relevance prediction combines the scores from the \textit{CM} and \textit{LR} stages as follows: $f(q,t)=\sigma(m(q,c), h(q,t))$, where $t\in c$ and $\sigma$ is a non-linear activation function that learns the weights to combine the cluster matching score ($m(q,c)$) and the relevance score within the cluster ($h(q,t)$). The top-$k$ types are then chosen based on $f(q,t)$.

\end{description}

\section{Type Clustering}
\label{sec:approach}

This section discusses the type clustering (\textit{TC}) step.
The main purpose of the TC phase is to reduce the label space for the BERT model used in the second stage (\textit{CM}) of the XBERT. More formally, the \textit{TC} step groups the set of types $T'$ from the training data into clusters $C$, where, $|C|<<|T|$.  In the past, question text-based TF-IDF vectors were used for clustering the types~\cite{Setty:2020:ISWCSMART}. This method does not use any external features for the type prediction. In this paper, we investigate how features derived from a KG, such as type similarity vectors and embeddings encoded using the type description text and KG structure, could improve the type prediction step and thereby end-to-end performance on the SMART task.

\subsection{Type Representation}

Embedding KG types allows us to use any clustering algorithm within the XBERT approach. More formally, each type is represented with a vector $\{ z_{t}:t \in T'\}$, where $z_{t} \in \mathbb{R}^d$ is a d-dimensional vector such that two KG types with high semantic similarity are closer together in the embedding space. Within the XBERT approach, these vectors are obtained by aggregating the features from the question text of the instances for a give type from the training data. For example, for the type \texttt{``dbo:SoccerPlayer''} features are aggregated from the related questions from the training data such as \emph{What are some players of England national football team?}, \emph{Which soccer players were born on Malta?} etc. In \cite{Setty:2020:ISWCSMART}, TF-IDF weights are used to represent the types. In this paper, we instead represent each type individually using textual and structural features from the type taxonomy of the underlying KG.
\subsubsection{KG Structure-Based Representations}
Based on the rationale that each type is defined by its neighborhood in the KG (for example, entities with that type and other related types) we propose the following two type representation methods:
\noindent    
\begin{description}[leftmargin=0.3cm]
    \item[Type similarity vector:]  Each type is represented by a vector of pairwise similarities to other types, based on the set of entities they share. We compute the similarity between two types $t$ and $t'$ using the Jaccard similarity, denoted as $J(t,t')$. 
    Each type $t$ then is represented as a vector of pairwise similarities between the type and all other types $t_1\dots t_n$ in the KG:
    \[
    z_{t}^{jaccard} = [J(t,t_1), J(t,t_2), \cdots, J(t,t_n)] \in \mathbb{R}^{|T|} ~.
    \]
        \item[RDF2Vec embeddings:] To leverage the neighborhood structure of KGs, we use network embedding representations.  
        Specifically, we use RDF2Vec~\cite{Ristoski:2016:ISWC} since it is  trained on both DBPedia and Wikidata, and it provides embeddings for KG types in addition to entities.  In short, RDF2Vec uses random walks to construct the sequences of nodes in the KGs.  These sequences are treated like sentences and distributed representation of the nodes in the RDF graphs are learned, similar to how Word2Vec models~\cite{Mikolov:2013:NIPS} are trained. We use the pre-trained 200-dimensional RDF2Vec SkipGram embeddings for both DBpedia and Wikidata.\footnote{\url{http://rdf2vec.org}}

    \end{description}

\noindent

\subsubsection{BERT-based Representations}

Next, we use textual descriptions of the types from the KG to represent them. These are sparse and short in practice, therefore, we augment them with the descriptions of entities with that type. A similar strategy has been used in retrieval methods for entity-bearing queries~\cite{Balog:2012:HTT}. Based on this we propose the following:

\begin{description}[leftmargin=0.3cm]
    \item[BERT type embeddings:] Each type is represented by concatenating the type description and the descriptions of the entities that are of that type. This text is then encoded using the BERT model to obtain an embedding for the type. More formally, a type description is a sequence of words $t_w$ and each entity description is a sequence of words $e^i_w$, that are concatenated with a special $[SEP]$ token and encoded as below:
   \begin{equation}
\small
\begin{aligned}
t_w &= [t_{w_1}, t_{w_2}, \ldots, t_{w_n}] \\
e^i_w &= [e^i_{w_1}, e^i_{w_2}, \ldots, e^i_{w_m}] \\
Desc_t &= [t_w[SEP]e^1_w[SEP]e^2_w[SEP]\ldots[SEP]e^l_w] \\
z_t^{BERT} &= BERT(Desc_t) \in \mathbb{R}^{d_{BERT}} ~,\\
\end{aligned}
\end{equation}
where $d_{BERT}$ is the BERT embedding dimension (1024). 

      \item[Fine-tuned BERT type embeddings:] The BERT model used above is pre-trained using task independent objectives such as masked language modeling and next sentence prediction. Alternatively, we can fine-tune the BERT model specifically for the cluster matching stage to encode the type embeddings. Intuitively, this would provide improved embeddings for the types.
\end{description}

\noindent
\subsection{Clustering Algorithm}
{Once we have the type representations, any clustering algorithm can be used to group the types into $|C|$ clusters. In this paper, we specifically report the results using simple a K-Means clustering algorithm. We note that other clustering methods were also tried (including KD-Trees and Spherical K-Means clustering), but the choice of clustering algorithm did not have a significant impact on type clustering performance.

\begin{table*}[h!]
\caption{Results for the type prediction and end-to-end SMART task. Best scores in each metric are boldfaced. $\dagger$ indicates statistically significant result with $p < 0.05$ and Bonferroni correction when compared with the baseline solution from \cite{Setty:2020:ISWCSMART}.}
\vspace{-0.5\baselineskip}
\label{tab:dbpedia1}
\centering
\begin{tabular}{P{3.8cm}l@{~~}l@{~~}l@{~~}l@{~~}l@{~~}c@{~~}c@{~~}c@{~~}ccc} 
    \toprule
       \multirow{3}{*}{\textbf{Method}} & & \multicolumn{7}{c}{\textbf{DBpedia}} & & \multicolumn{2}{c}{\textbf{Wikidata}} \\
    & & \multicolumn{3}{c}{\textbf{Type prediction}} & & \multicolumn{3}{c}{\textbf{End-to-end}}  & &  \textbf{Type prediction} & \textbf{End-to-end}\\
    \cline{3-5} \cline{7-9} \cline{11-12} 
    & & \textbf{NDCG@3} & \textbf{NDCG@5} & \textbf{NDCG@10} & & \textbf{NDCG@3} & \textbf{NDCG@5} & \textbf{NDCG@10} && \textbf{MRR} & \textbf{MRR}\\
    \midrule
    Question text (TF-IDF) & & 0.717 & 0.693 & 0.650 & & 0.824 & 0.811 & 0.787 && 0.66 & 0.76\\
    KG-TypeSim  & & 0.725 & 0.697  & 0.662 && 0.828 & 0.813 &  0.793 & & 0.67 & 0.77\\
    KG-RDF2Vec & & 0.729 & 0.701 & 0.656 && 0.831 & 0.815 &  0.791 & & 0.67 & 0.78 \\
    BERT-TypeDesc & & 0.727 & 0.706 & 0.665 && 0.830 & 0.818 & 0.795 && 0.67 &  0.78 \\
    BERT-TypeDesc~(fine-tuned) & & \textbf{0.734$^\dagger$} & \textbf{0.712$^\dagger$} & \textbf{0.678$^\dagger$} && \textbf{0.834} & \textbf{0.822} & \textbf{0.802} & & \textbf{0.68} & \textbf{0.79} \\
    \bottomrule
    \end{tabular}
    
\end{table*}
\section{Experiments}
\label{sec:exps}

In this section, we evaluate the different type representations for the \textit{TC} stage and end-to-end SMART task.
\subsection{Experimental Setup}
\label{subsec:setup}

\subsubsection{Datasets}
\label{subsec:data}

The DBpedia and Wikidata datasets each have around 17-18k training instances and 4k test instances.
In both datasets, the majority of questions belong to the \texttt{Resource} category and have one or more ground truth types from the respective KG. For more details, see~\cite{Mihindukulasooriya:2020:ISWCSMART}. 

\subsubsection{Evaluation Metrics}
We follow the evaluation method used in the official SMART challenge.
Type prediction is cast as a ranking task and is evaluated using rank-based metrics. It, however, considers only those questions that fall into the
\texttt{literal} (\texttt{number}, \texttt{string}, \texttt{date}) or \texttt{resource} answer categories. 
Furthermore, evaluation is performed differently for DBpedia and for Wikidata, given the nature of their respective type taxonomies.
Types in the DBpedia Ontology are organized hierarchically, up to 7 levels deep.  There, a graded evaluation metric, Normalized Discounted Cumulative Gain (NDCG@k), is used. Specifically, literal answer types are either correct or incorrect, while resource answer types receive gain values depending on their distance from the gold types in the type hierarchy~\cite{Balog:2012:HTT}. In case of Wikidata, the type hierarchy is rather flat. Therefore, type prediction is evaluated using a binary notion of relevance, with Mean Reciprocal Rank (MRR) as the metric.

\subsubsection{Implementation and baseline}
We compute the KG type features using the DBpedia dump from October 2016 %
and Wikidata  dump from May 19, 2021. %
We use the RoBERTa~\cite{Liu:2019:arXiv} (roberta-large\footnote{\url{https://huggingface.co/roberta-large}}) for both encoding the textual descriptions and for fine-tuning the classifier in the second stage (\textit{CM}). We ran all experiments on a Linux server with Nvidia Tesla V100 GPU with 32 GB memory. Note that the evaluation script provided by the SMART task organizers had a known bug\footnote{\url{https://github.com/smart-task/smart-dataset/issues/10}}, 
which we fixed for this paper. Therefore, the scores we report for the baseline method in Table~\ref{tab:dbpedia1} slightly are higher than those reported in \cite{Setty:2020:ISWCSMART}.

We use the Question text (TF-IDF) method~\cite{Setty:2020:ISWCSMART} as baseline in the experiments. We represent our KG derived features as (1) \textbf{KG-TypeSim}: pairwise KG-TypeSim vector, (2) \textbf{RDF2Vec}: RDF2Vec embeddings, (3) \textbf{BERT-TypeDesc}: Type description (TD) encoded with BERT and (4) \textbf{BERT-TypeDesc (fine-tuned)}: TD encoded with fine-tuned BERT. 

\subsubsection{Parameter settings}

The number of clusters ($|C|$) is a hyperparameter of the \emph{TC} stage, which we set experimentally, using cross-validation on the training set.  We consider the values $\{32, 64, 128, 256, 512\}$.
We observe the best performance for DBpedia with 64 clusters and for Wikidata with 128 clusters, which will be the values used in the remaining of the experiments.
We also note that this parameter has hardly any effect in case of Wikidata.

\subsection{Results}
\label{sec:results}

Table~\ref{tab:dbpedia1} shows the results for type classification (i.e., only for the \texttt{resource} category) as well as on the end-to-end task (which also includes coarse category classification).

\begin{description}
\item[RQ1] Can we improve the clustering stage for the SMART task by using additional signals from the KG?
\end{description}

\noindent
We observe that the TF-IDF-based question text baseline performs reasonably well. The KG features ``KG-TypeSim'' and ``KG-RDF2Vec'' yield slightly higher NDCG scores. The best performance is obtained with the BERT models that encode textual descriptions of types as well as of corresponding entities from the KG. We also find that fine-tuning can further enhance performance. 

We also observe a similar pattern regarding end-to-end performance improvements. However, the results are not statistically significant---this is because the end-to-end pipeline also includes coarse category classification. Specifically, end-to-end evaluation also includes the questions which belong to the \texttt{boolean} and various \texttt{literal} categories, and hence downplay the overall impact of fine-grained type classification.

\begin{description}
 \item[RQ2] How well do these findings generalize across KGs, which differ in the characteristics of their type systems?
\end{description}

\noindent
When we compare the performance of DBpedia and Wikidata, we see notable (and in some cases statistically significant) improvements for DBpedia, while Wikidata shows only marginal differences.  This is also apparent from the fact that the results are statistically significant for DBpedia but not for the Wikidata. We attribute this to the fact the DBpedia has a rich type system, which can by leveraged for type prediction, while Wikidata has a relatively flat type hierarchy.

\subsection{Analysis}
In this section, we analyze anecdotal examples to highlight the strengths and weaknesses of our approach. 

For types with good descriptions from KG, \textbf{BERT-TypeDesc} methods perform well. For example, ``Which are exclusions of Asperger syndrome?'', the ground-truth is \texttt{`dbo:Disease'} and our method predicts \texttt{[`dbo:Biomolecule', `dbo:Gene', `dbo:Disease']} as top-3 types. While, the baseline method predicts \texttt{['dbo:Media', 'dbo:EthnicGroup', 'dbo:Religious']}, since the question text alone is insufficient. Our methods fail when the type descriptions and questions have little in common. For example, for ``Name a participant of the American Revolutionary War.'' our method predicts
\texttt{[`dbo:Agent', `dbo:Person', `dbo:AmericanFootballTeam']} while the ground-truth is  \texttt{[`dbo:Country', `dbo:State']}.

In some cases method is able to predict more appropriate types, while the ground-truth is incorrect. For example, ``Which is the vessel class of the galleon'' has the ground-truth \texttt{`dbo:Work'} and our method predicts \texttt{[`dbo:MeanOfTransportation', `dbo:Ship', `dbo:Spacecraft']}.

\section{Conclusion and Future Work}
\label{sec:concl}
In this paper, we have proposed different ways to represent KG types to improve the clustering stage of the XBERT pipeline for the SMART task. We have shown that features derived from KGs to represent types are beneficial, especially for KGs with a rich type system, such as DBpedia. Our error analysis suggests that the BERT-based type embeddings are effective in capturing question context, and our method is able to deal with noisy training data.

 In addition to improving type prediction, cluster-based features could also be used for improving the third stage of the XBERT pipeline for SMART in the future. 
 It would also be worth exploring if KGs with a richer structure, like DBpedia, can be used to improve prediction effectiveness on KGs with relatively flat type systems, like Wikidata.
 Moreover, improved type clustering can potentially be useful for other tasks as well that involve KGs. Specifically, in future work, we would like to explore how to use answer type prediction in a conversational setting. For instance, this method can be used to decide if a question can be answered with a KG.

\section{Acknowledgements}
I would like to thank Prof. Krisztian Balog for his valuable input and contributions for this paper.
\balance
\newpage

\balance
\bibliographystyle{abbrv}
\bibliography{jcdl2023}

\begin{thebibliography}{10}

\bibitem{Balog:2018:book}
K.~Balog.
\newblock {\em Entity-Oriented Search}, volume~39 of {\em The Information
  Retrieval Series}.
\newblock Springer, 2018.

\bibitem{Balog:2012:HTT}
K.~Balog and R.~Neumayer.
\newblock Hierarchical target type identification for entity-oriented queries.
\newblock In {\em Proceedings of the 21st ACM international conference on
  Information and knowledge management}, CIKM '12, pages 2391--2394, 2012.

\bibitem{Bast:2015:CIKM}
H.~Bast and E.~Haussmann.
\newblock More accurate question answering on freebase.
\newblock In {\em Proceedings of the 24th ACM International on Conference on
  Information and Knowledge Management}, CIKM '15, pages 1431--1440, 2015.

\bibitem{Bill:2020:ISWCSMART}
E.~Bill and E.~Jim{\'e}nez-Ruiz.
\newblock Question embeddings for semantic answer type prediction.
\newblock In {\em Proceedings of the SeMantic AnsweR Type prediction task
  {(SMART)} at {ISWC} 2020}, ISWC SMART '20, pages 71--80, 2020.

\bibitem{Chang:2020:KDD}
W.-C. Chang, H.-F. Yu, K.~Zhong, Y.~Yang, and I.~S. Dhillon.
\newblock Taming pretrained transformers for extreme multi-label text
  classification.
\newblock In {\em Proceedings of the 26th ACM SIGKDD International Conference
  on Knowledge Discovery \& Data Mining}, KDD '20, pages 3163--3171, 2020.

\bibitem{Ferrucci:2010:AI}
D.~A. Ferrucci, E.~W. Brown, J.~Chu{-}Carroll, J.~Fan, D.~Gondek, A.~Kalyanpur,
  A.~Lally, J.~W. Murdock, E.~Nyberg, J.~M. Prager, N.~Schlaefer, and C.~A.
  Welty.
\newblock Building {Watson}: {A}n overview of the {DeepQA} project.
\newblock {\em {AI} Magazine}, 31(3):59--79, 2010.

\bibitem{Garigliotti:2017:SIGIR}
D.~Garigliotti, F.~Hasibi, and K.~Balog.
\newblock Target type identification for entity-bearing queries.
\newblock In {\em Proceedings of the 40th International ACM SIGIR Conference on
  Research and Development in Information Retrieval}, SIGIR '17, pages
  845--848, 2017.

\bibitem{Garigliotti:2019:IET}
D.~Garigliotti, F.~Hasibi, and K.~Balog.
\newblock Identifying and exploiting target entity type information for ad hoc
  entity retrieval.
\newblock {\em Information Retrieval Journal}, 22(3):285--323, 2019.

\bibitem{Huang:2008:EMNLP}
Z.~Huang, M.~Thint, and Z.~Qin.
\newblock Question classification using head words and their hypernyms.
\newblock In {\em Proceedings of the 2008 Conference on empirical methods in
  natural language processing}, EMNLP '08, pages 927--936, 2008.

\bibitem{Kamath:2019:CS}
S.~Kamath, B.~Grau, and Y.~Ma.
\newblock Predicting and integrating expected answer types into a simple
  recurrent neural network model for answer sentence selection.
\newblock {\em Computaci{\'o}n y Sistemas}, 23(3):665--673, 2019.

\bibitem{Kwok:2001:WWW}
C.~C. Kwok, O.~Etzioni, and D.~S. Weld.
\newblock Scaling question answering to the web.
\newblock In {\em Proceedings of the 10th international conference on World
  Wide Web}, WWW '01, pages 150--161, 2001.

\bibitem{Li:2006:NLE}
X.~Li and D.~Roth.
\newblock Learning question classifiers: the role of semantic information.
\newblock {\em Natural Language Engineering}, 12(3):229--249, 2006.

\bibitem{Liu:2020:arXiv}
W.~Liu, H.~Wang, X.~Shen, and I.~Tsang.
\newblock The emerging trends of multi-label learning.
\newblock 2020.

\bibitem{Liu:2019:arXiv}
Y.~Liu, M.~Ott, N.~Goyal, J.~Du, M.~Joshi, D.~Chen, O.~Levy, M.~Lewis,
  L.~Zettlemoyer, and V.~Stoyanov.
\newblock Roberta: A robustly optimized bert pretraining approach.
\newblock 2019.

\bibitem{Mihindukulasooriya:2020:ISWCSMART}
N.~Mihindukulasooriya, M.~Dubey, A.~Gliozzo, J.~Lehmann, A.~N. Ngomo, and
  R.~Usbeck, editors.
\newblock {\em Proceedings of the SeMantic AnsweR Type prediction task
  {(SMART)} at {ISWC} 2020 Semantic Web Challenge co-located with the 19th
  International Semantic Web Conference {(ISWC} 2020), Virtual Conference,
  November 5th, 2020}, volume 2774 of {\em {CEUR} Workshop Proceedings}.
  CEUR-WS.org, 2020.

\bibitem{Mikolov:2013:NIPS}
T.~Mikolov, I.~Sutskever, K.~Chen, G.~S. Corrado, and J.~Dean.
\newblock Distributed representations of words and phrases and their
  compositionality.
\newblock {\em Advances in neural information processing systems},
  26:3111--3119, 2013.

\bibitem{Moldovan:2003:TOIS}
D.~Moldovan, M.~Pa\c{s}ca, S.~Harabagiu, and M.~Surdeanu.
\newblock Performance issues and error analysis in an open-domain question
  answering system.
\newblock {\em ACM Trans. Inf. Syst.}, 21(2):133--154, apr 2003.

\bibitem{Nikas:2020:ISWCSMART}
C.~Nikas, P.~Fafalios, and Y.~Tzitzikas.
\newblock Two-stage semantic answer type prediction for question answering
  using {BERT} and class-specificity rewarding.
\newblock In {\em Proceedings of the SeMantic AnsweR Type prediction task
  {(SMART)} at {ISWC} 2020}, SMART '20, pages 19--28, 2020.

\bibitem{Nikas:2021:ISWC}
C.~Nikas, P.~Fafalios, and Y.~Tzitzikas.
\newblock Open domain question answering over knowledge graphs using keyword
  search, answer type prediction, sparql and pre-trained neural models.
\newblock In {\em International Semantic Web Conference}, pages 235--251, 2021.

\bibitem{Perevalov:2021:ISWCSMART}
A.~Perevalov and A.~Both.
\newblock Augmentation-based answer type classification of the smart dataset.
\newblock In {\em Proceedings of the SeMantic AnsweR Type prediction task
  {(SMART)} at {ISWC} 2020}, ISWC SMART '20, pages 1--9, 2020.

\bibitem{Perevalov:2021:ISWC}
A.~Perevalov and A.~Both.
\newblock Improving answer type classification quality through combined
  question answering datasets.
\newblock In {\em International Conference on Knowledge Science, Engineering
  and Management}, ISWC '21, pages 191--204, 2021.

\bibitem{Prabhu:2018:WWW}
Y.~Prabhu, A.~Kag, S.~Harsola, R.~Agrawal, and M.~Varma.
\newblock Parabel: Partitioned label trees for extreme classification with
  application to dynamic search advertising.
\newblock In {\em Proceedings of the 2018 World Wide Web Conference}, WWW '18,
  pages 993--1002, 2018.

\bibitem{Riloff:2000:ANLP-NAACL}
E.~Riloff and M.~Thelen.
\newblock A rule-based question answering system for reading comprehension
  tests.
\newblock In {\em ANLP-NAACL 2000 Workshop: Reading Comprehension Tests as
  Evaluation for Computer-Based Language Understanding Systems}, ANLP-NAACL
  '00, pages 13--19, 2000.

\bibitem{Ristoski:2016:ISWC}
P.~Ristoski and H.~Paulheim.
\newblock Rdf2vec: {RDF} graph embeddings for data mining.
\newblock In {\em The Semantic Web - {ISWC} 2016 - 15th International Semantic
  Web Conference, Kobe, Japan, October 17-21, 2016, Proceedings, Part {I}},
  volume 9981 of {\em ISWC '16}, pages 498--514, 2016.

\bibitem{Roy:2021:QABook}
R.~S. Roy and A.~Anand.
\newblock {\em Question Answering for the Curated Web: Tasks and Methods in QA
  over Knowledge Bases and Text Collections}.
\newblock Morgan \& Claypool Publishers, 2021.

\bibitem{Setty:2020:ISWCSMART}
V.~Setty and K.~Balog.
\newblock Semantic answer type prediction using {BERT} {IAI} at the {ISWC}
  {SMART} task 2020.
\newblock In {\em Proceedings of the SeMantic AnsweR Type prediction task
  {(SMART)} at {ISWC} 2020}, ISWC SMART '20, pages 10--18, 2020.

\bibitem{Shen:2019:EMNLP}
T.~Shen, X.~Geng, Q.~Tao, D.~Guo, D.~Tang, N.~Duan, G.~Long, and D.~Jiang.
\newblock Multi-task learning for conversational question answering over a
  large-scale knowledge base.
\newblock In {\em Proceedings of the 2019 Conference on Empirical Methods in
  Natural Language Processing and the 9th International Joint Conference on
  Natural Language Processing (EMNLP-IJCNLP)}, EMNLP '19, pages 2442--2451,
  2019.

\bibitem{Sun:2015:WWW}
H.~Sun, H.~Ma, W.-t. Yih, C.-T. Tsai, J.~Liu, and M.-W. Chang.
\newblock Open domain question answering via semantic enrichment.
\newblock In {\em Proceedings of the 24th International Conference on World
  Wide Web}, WWW '15, pages 1045--1055, 2015.

\bibitem{Vallet:2008:SIGIR}
D.~Vallet and H.~Zaragoza.
\newblock Inferring the most important types of a query: A semantic approach.
\newblock In {\em Proceedings of the 31st Annual International ACM SIGIR
  Conference on Research and Development in Information Retrieval}, SIGIR '08,
  pages 857--858, 2008.

\bibitem{Vallurupalli:2020:ISWCSSMART}
S.~Vallurupalli, J.~Sleeman, and T.~Finin.
\newblock Fine and ultra-fine entity type embeddings for question answering.
\newblock In {\em Proceedings of the SeMantic AnsweR Type prediction task
  {(SMART)} at {ISWC} 2020}, ISWC SMART '20, pages 57--70, 2020.

\bibitem{Voorhees:2001:TREC}
E.~M. Voorhees.
\newblock The trec question answering track.
\newblock {\em Nat. Lang. Eng.}, 7(4):361–378, Dec. 2001.

\bibitem{Yavuz:2016:EMNLP}
S.~Yavuz, I.~G{\"u}r, Y.~Su, M.~Srivatsa, and X.~Yan.
\newblock Improving semantic parsing via answer type inference.
\newblock In {\em Proceedings of the 2016 Conference on Empirical Methods in
  Natural Language Processing}, EMNLP '16, pages 149--159, 2016.

\bibitem{Yu:2020:arXiv}
H.-F. Yu, K.~Zhong, and I.~S. Dhillon.
\newblock Pecos: Prediction for enormous and correlated output spaces.
\newblock 2020.

\end{thebibliography}

\end{document}